# EVIDENCE FEED FORWARD HIDDEN MARKOV MODEL: A NEW TYPE OF HIDDEN MARKOV MODEL

Michael Del Rose[1], Christian Wagner[2], and Philip Frederick[1]

[1] U.S. Army Tank Automotive Research, Development, and Engineering Center, Warren, Michigan, U.S.A

[2] Department of Industrial and Systems Engineering, Oakland University, Rochester, Michigan, U.S.A

**ABSTRACT**

*The ability to predict the intentions of people based solely on their visual actions is a skill only performed by humans and animals. The intelligence of current computer algorithms has not reached this level of complexity, but there are several research efforts that are working towards it. With the number of classification algorithms available, it is hard to determine which algorithm works best for a particular situation. In classification of visual human intent data, Hidden Markov Models (HMM), and their variants, are leading candidates.*

*The inability of HMMs to provide a probability in the observation to observation linkages is a big downfall in this classification technique. If a person is visually identifying an action of another person, they monitor patterns in the observations. By estimating the next observation, people have the ability to summarize the actions, and thus determine, with pretty good accuracy, the intention of the person performing the action. These visual cues and linkages are important in creating intelligent algorithms for determining human actions based on visual observations.*

*The Evidence Feed Forward Hidden Markov Model is a newly developed algorithm which provides observation to observation linkages. The following research addresses the theory behind Evidence Feed Forward HMMs, provides mathematical proofs of their learning of these parameters to optimize the likelihood of observations with a Evidence Feed Forwards HMM, which is important in all computational intelligence algorithm, and gives comparative examples with standard HMMs in classification of both visual action data and measurement data; thus providing a strong base for Evidence Feed Forward HMMs in classification of many types of problems.*

## 1. INTRODUCTION

Visual Understanding (VU) is the ability for a machine to understand events, items, and scenarios through visual cues, or visual data. It is a very important and complex process used in many artificial and computational intelligence research programs. The need for VU is increasing with the growing advances in technology that require VU algorithms to be taken out of the research labs and into fully developed programs [11, 12, 13, 32, 83].

A sub research area of VU is Visual Human Intent Analysis (VHIA). This area may also be referred to as *visual human behavior identification*, *action or activity recognition*, and *understanding human actions from visual cues*. VHIA concentrates on the visual identification of actions made by a human. There are many different names associated with VHIA that describe the specific process. In static self security systems *visual human behavior identification* systems will aide or replace security guards monitoring CCTV feeds [13]. Television stations and the gaming community will require *activity recognition* systems to automatically categorize and store or quickly search for certain scenes in a database [12]. The military is pushing robotics to replace the soldier, thus requiring the need to *understand human actions from visual cues* to determine hostile actions from people so the robot can take appropriate actions to secure itself [83]. These are just a few names of the many names for VHIA.

Evidence Feed Forward Hidden Markov Models are designed to better handle many of the shortcomings not addressed in current classification systems. The motivation of this new research is to provide a way of better detecting human movements for classification of the person's activity. This classification will be based on the observations being linked, not just linking observations with events as described in standard Hidden Markov Models. Moving from better classifications in visual human activity to other types of non-visual data will be discussed. Classification results of Evidence Feed Forward Hidden Markov Models compared with results of standard Hidden Markov Models will also be shown for both visual and non-visual data. The goal of this research is to develop a more robust classification system then current standard Hidden Markov Models and to identify a new way of looking at the links between evidences from the classification system; in the case of Evidence Feed Forward HMMs this is described as observation to observation links.

The first section of this paper is the introduction. The second section is a brief overview of current research in the area of VHIA. The third section describes the Evidence Feed Forward theory and derives the equations for the three common problems of HMMs. The fourth and fifth sections use the Evidence Feed Forward HMM developed algorithms and apply them to two examples, one measurement based and one visual based. The final section is the conclusion.

## 2. CURRENT RESEARCH IN THE AREA OF VISUAL HUMAN INTENT ANALYSIS

The current research in the area of VHIA is split into six sub-sections which best describes the methods based on the volume of work: Non-traditional artificial intelligent (AI) methods, traditional AI methods, Markov models and Bayesian networks, grammars, traditional hidden Markov models (HMM), and non-traditional HMMs.

Non-traditional AI methods general do not have learning in them and rely on heavy processing of the data to determine the intent of the person. M. Cristani et al [1] uses non-traditional AI methods by taking in both audio and visual data to determine simple events in an office. First they remove foreground objects and segment the images in the sequence. This output is coupled with the audio data and a threshold detection process is used to identify unusual events. These event sequences are put into an audio visual concurrence matrix (AVC) to compare with known AVC events.

Template matching is performed by M. Dimitrijevic et. al. [2]. They developed a template database of actions based on five male and three female people. Each human action is represented by three frames of their 2D silhouette at different stages of the activity: the frame when the person first touches the ground with one of his/her feet, the frame at the midstride of the step, and the end frame when the person finishes touching the ground with the same foot. The three frame sets were taken from seven camera positions. When determining the event, they use a modified Chamfer's distance calculation to match to the template sequences in the database.

Traditional AI methods usually have some type of learning, either with known or unknown outcomes. Typical methods would include neural networks, fuzzy systems, and other common AI techniques. H. Stern et al. [3] created a prototype fuzzy system for picture understanding of surveillance cameras. His model is split into three parts, pre-processing module, a static object fuzzy system module, and a dynamic temporal fuzzy system module. The static fuzzy system module classifies pre-processed data as a single person, two people, three people, many people, or no people. The dynamic fuzzy system determines the intent of the person based on the temporal movements.

Grammars are often used to describe the sequence of events, usually through a type of words or characters. These words are developed by extrapolating the action to common words, which grouped together may make a sentence describing the action (in a non-literal sense). A. Ogale et. al. [4] uses probabilistic context free grammars (PCFG) in short action sequences of a person from video. Body poses are stored as silhouettes which are used in the construction of the PCFG. Pairs of frames are constructed based on their time slot: the pose from frame 1 and 2 are paired, the pose from frame 2 and 3 are paired, and so on. These pairs construct the PCFG for the given action. When testing the algorithm, the same procedure is followed. Comparing the testing data with the trained data is accomplished through Bayes: $P(s_k|p_i) = P(p_i|s_k)P(s_k)/P(p_i)$, where $s_k$ is the $k^{th}$ silhouette and $p_i$ is the $i^{th}$ pose.

There are a number of traditional and non-traditional Hidden Markov Models (HMM) that are used in trying to understand peoples actions based on visual sequences. Traditional HMMs are used to classify items that have a flow to them. This flow, generally related to time, determines the specific body pose of the action sequence and the HMM determines how likely the flow is. For non-traditional HMMs, they are just an offshoot to the standard HMM except they add another facet which provides better results based on the type of data coming in, generally used in specific examples. A few include Yamato et. al. [5] used HMMs to recognize six tennis strokes with a 25x25 mess feature matrix to describe body positions in each frame. Wilson and Bobick [6] use a Parametric Hidden Markov Model (PHMM) to recognize hand gesture. Oliver et. al. [7] developed a method to detect and classify interactions between people using a Coupled Hidden Markov Model (CHMM) based on simulations. Multi-Observation Hidden Markov Models (MOHMM) are discussed in both [8] and [9] from Xaing and Gong for recognizing break points in video content for separation of activities and detect piggybacking of peoples going through a security door, respectively.

## 3. Evidence Feed Forward Hidden Markov Models

### 3.1 Evidence Feed Forward Hidden Markov Model Introduction

The Evidence Feed Forward Hidden Markov Model is more than an extension or a variance of Hidden Markov Models (HMM), like Parametric HMMs or Hierarchical HMMs, because it links the current observations to the next observation in the sequence, assigns a probability associated with this, and has the ability to better describe visual actions. The Evidence Feed Forward HMM gives the impression of disregarding the rules of causality, as suggested by HMM theory, by providing the link between observations that is not through the hidden nodes, which in the strict sense, Markov models current state only depends on the previous states.

The assumption of breaking causality is relatively true in standard HMMs, but if you look at the reason for making the model is to "model the event" not create real world, then we can relax the causality rule, at least how we look at causality in modeling with HMMs. It is also important to note that in many cases, previous events will affect future events. Take, for example a common weather example found in many textbooks: A person is locked inside a windowless building and would like to know whether it is raining or not outside. The only evidence he has is the observation of his boss coming inside with or without an umbrella. He constructs an HMM to make his decision. Figure 1 shows the hidden layer is represented by the nodes, Rain (R) and No Rain (NR). The observation is represented by the nodes, Umbrella (U) and No Umbrella (NU). This example shows that the evidence (observation) is only dependent on the hidden layer and not vice versa. Also, the hidden layer is only dependent on the previous day's weather (with some probability). HMMs and their variants do not take into account the probability of the current observation's effect on the next observation. Obviously, in our example, current observation of an umbrella (or no observation of an umbrella) does not affect tomorrow's observation of an umbrella (or no umbrella). However, if you look at the observation as another event carried out by the boss, then you can see that there may be some probability in the HMM that has been overlooked when modeling this example. These two events can be tied together in this model; the event of the boss choosing an umbrella and the event of the person seeing the umbrella. Independence with respect to the model is still intact. Causality of the observation is looked at differently than in a standard HMM model. The underlying reason the observations have an effect on the next observation is based on the event by the boss and not looked at as just an observation. Thus, causality is still adhered to. Specifically, if, for example, the boss comes into the building without an umbrella and it is raining then one can probably assume that the boss may be more likely to carry an umbrella the next day since he did not like getting wet. The same may be true if he did carry an umbrella. This connects the evidence of each event to the evidence of the next nodes event as shown in Figure 2.

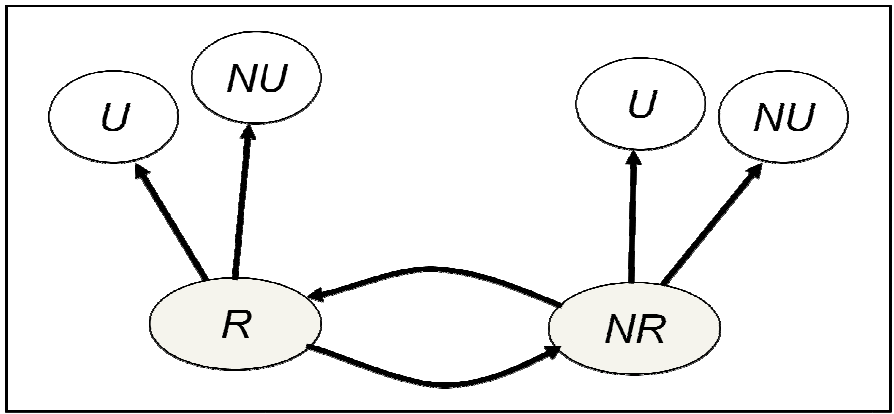

Figure 1: Weather example using HMMs. The nodes in the hidden layer are represented by Rain (R) and No Rain (NR). The nodes in the observation layer are represented by U and NU.

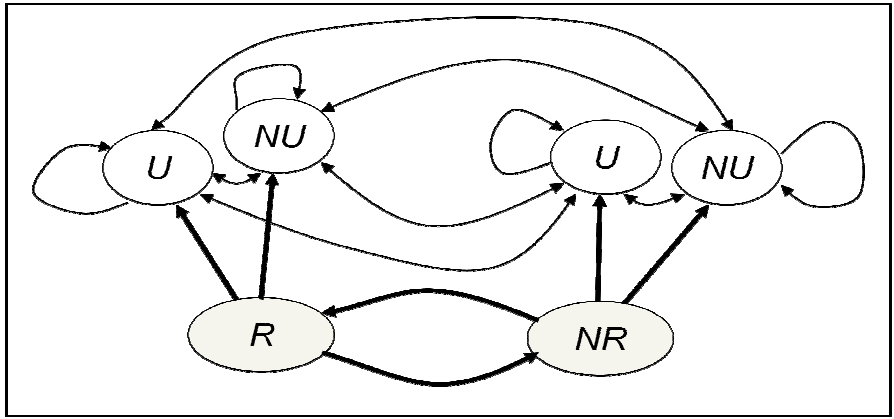

Figure 2: Weather example using Evidence Feed-Forward HMMs. Linkages between the evidence (marked U and NU) and the other evidences represent the probability of the evidence occurring based on the previous evidence.

In this example, Figure 2 shows that the edges connecting observations to observations are represented as bi-directional edges which may suggest that the probability is the same in either direction. This is not the case. In fact, it can be shown that the problem can be reduced in complexity so that the probabilities of the edges connecting the evidence rely only on the evidence and the current hidden state and not on the future hidden state. The bi-directional edges are illustrated as such due to simplification of the drawing. In this weather example, assume that the boss forgets his umbrella while it is raining outside. One can assume that the probability associated with the boss carrying his umbrella the next day should be greater since he did not like the feeling of getting wet. He probably wants to reduce his chance of getting wet, thus puts more emphasis on carrying an umbrella whether it is raining outside or not. It can be assumed that the probability represented in the edge from U given R to U given NR has the same probability as the edge going to U given R. In a sense, one can think of this as an addition to the probabilities based on the person (or thing) which the outcome (observation) is directly related to.

### 3.2 Evidence Feed Forward Hidden Markov model derivation

Before jumping into deriving equations, the variables should formally be defined. Let N be the total number of hidden nodes and M be the total number of observations. Let T be the total number of transitions (or time). Let $Q_t$ be the current state at time t, $1 \leq t \leq T$; $Q_t = S_i$ means the state at time t is $S_i$ where $1 \leq i \leq N$. The observations at time t are represented by $O_t$. $O_t = V_h$ means the observation at time t is $V_h$ where $1 \leq h \leq M$.

The probability of transitioning from one hidden state to another is captured by $a_{ij}$, where i is the current state and j is the next state, $1 \leq i,j \leq N$. Let A be an NxN matrix which captures all the $a_{ij}$'s. The rows represent the current state (i) and the columns represent the next state (j). If there are three hidden nodes in the Evidence Feed Forward HMM, N=3, then A would be:

$$A = \begin{bmatrix} a11 & a12 & a13 \\ a21 & a22 & a32 \\ a31 & a32 & a33 \end{bmatrix}$$

Let $b_{jh}$ be the probability of observing $V_h$ while in state $S_j$. Let B be an NxM matrix which holds the observation probabilities. If there were three hidden layers and two observations, N=3 and M=2, then B would be

$$B = \begin{bmatrix} b11 & b12 \\ b21 & b22 \\ b31 & b32 \end{bmatrix}$$

Let $c_i(h,k)$ be the probability of being in state $S_i$ and currently observing $V_h$ with the next observation being $V_k$ for $1 \leq h,k \leq M$. Let C be the MxMxN matrix represented as shown below

$$C_1 = \begin{bmatrix} c_1(1,1) & c_1(1,2) \\ c_1(2,1) & c_1(2,2) \end{bmatrix}, C_2 = \begin{bmatrix} c_2(1,1) & c_2(1,2) \\ c_2(2,1) & c_2(2,2) \end{bmatrix}, \text{and } C_3 = \begin{bmatrix} c_3(1,1) & c_3(1,2) \\ c_3(2,1) & c_3(2,2) \end{bmatrix}.$$

Let the probabilities of starting in state $S_i$ be represented as $\pi_i$. $\pi$ is a 1xN vector which holds all the $\pi_i$ values. $\pi = [\pi_1 \quad \pi_2 \quad \pi_3]$ for a three hidden node Evidence Feed-Forward HMM.

Similar to the standard HMM, the Evidence Feed Forward HMM is represented by $\lambda$ but with an added parameter. $\lambda$ has the parameters A, B, C, and $\pi$ and can be written as $\lambda=(\pi, A, B, C)$. Either $\lambda$ or $(\pi, A, B, C)$ can represent the Evidence Feed Forward HMM. Given the model $\lambda$ at time t in the current state $S_i$ observing $V_h$, what is the probability of being in state $S_j$ observing $V_k$? The calculation of transitioning from state $S_i$ to state $S_j$ is $a_{ij}$, observing $V_h$ from state $S_j$ is $b_{jh}$, and being in state $S_i$ observing $V_h$ with next observation being $V_k$ is $c_i(h,k)$. This probability is $a_{ij} \cdot b_{jh} \cdot c_i(h,k)$. When possible, to increase the readability, $S_i$ for all i will be represented as just i. Also, $V_h$ for all h will be represented by just h.

There are three typical problems that all HMMs should solve:
1. What is the probability of the observation sequence $O = (O_1,O_2,…,O_T)$ given the model $\lambda$? This is asking to find $P(O|\lambda)$.
2. What is the most optimal hidden state path given the observations O and the model $\lambda$? This is asking to solve $P(Q|O,\lambda)$ where $Q = (Q_1,Q_2,…,Q_T)$.
3. Given a number of observations, what is the optimal parameters of $\lambda$ which maximizes $P(O|\lambda)$? This is the learning problem.

For a detailed tutorial on how standard HMMs solve these problems the reader is referred to Rabiner [13].

To solve the first problem, "What is the probability of the observation sequence $O = (O_1,O_2,…,O_T)$ given the model $\lambda$?", a forward algorithm procedure is developed to compute $\alpha_i(t) = p(O_1,O_2,…O_t,Q_t = i|\lambda)$. When t = T, $P(O|\lambda)$ is found by summing all the $\alpha_i$'s at time T. The forward algorithm procedure is:
1. $\alpha_i(1) = \pi_i b_i(O_1)$ for all i, $0 \leq i$, $t \leq T$, and $b_i(O_1) = b_{ih}$ for some h which $O_1 = V_h$. Notice that there is no c term because this is the initial starting state calculation for $\alpha_i$ so there is no observation to observation in the calculation of the initial probabilities.
2. $\alpha_j(t+1) = \left[\sum_{i=1}^{N} \alpha_i(t) a_{ij} c_i(O_t, O_{t+1})\right] b_j(O_{t+1})$, where $c_i(O_t, O_{t+1})$ is $c_i(h,k)$ for $O_t = V_h$ and $O_{t+1} = V_k$ and N is the total number of hidden states.
3. $P(O|\lambda) = \sum_{i=1}^{N} \alpha_i(T)$.

From the final part of step 2 where $\alpha_i(T) = P(O_1,O_2,…,O_T,Q_T = i|\lambda)$, we find the probability of the observation sequence and the final state $Q_T = i$ given the model. By summing up all the $\alpha_i$'s we get the final probability of $P(O|\lambda)$ shown in step three.

A backwards algorithm procedure can also be developed to find $P(O|\lambda)$. In the backwards algorithm, the starting state is at time T and the algorithm is worked backwards towards time t. The variable $\beta_i$ is defined as $\beta_i(t) = P(O_{t+1},O_{t+2},…,O_T | Q_t = i, \lambda)$.

1. $\beta_i(T) = 1$
2. $\beta_i(t) = \left[\sum_{j=1}^{N} a_{ij} b_j(O_{t+1}) \beta_j(t+1)\right] c_i(O_t, O_{t+1})$
3. $P(O|\lambda) = \sum_{i=1}^{N} \beta_i(1) \pi_i b_i(O_1)$.

It should be noted that the probability of the observations given the model using both the forward and backwards algorithms are used later to help find answers to the remaining two Evidence Feed-Forward HMM problems.

The second common HMM problem which the Evidence Feed-Forward HMM should solve is computing the optimal path of hidden states from the observations given the model. Optimal path could mean many different things. One way to look at optimal paths is to find the maximum probability moving from one state/observation to another, considering only the path which is maximum for only the current transition. This has the possibility of transitioning to a state which leaving from is impossible.

In this research, finding the optimal path is considered the same as finding the maximum probability for the entire series and not individual maximums. Optimal path is assumed that one is looking for the path that gives the maximum probability of the state sequence given the observations and the model; maximizing $P(Q|O,\lambda)$. This solution requires the use of both the forward and backwards algorithms. To do this two new variables must be created, $\delta$ and Path. $\delta$ is defined as the running probability of paths at time t. Path is the current path found from computing $\delta$.

1. $\delta_1(i) = \pi_i b_i(O_1)$. Path = [].
2. $\delta_t(j) = \max_{1 \leq i \leq n}[\delta_{t-1} a_{ij} b_j(O_t) c_i(O_{t-1}, O_t)]$. Path is the state which this is maximized. Add the state to the Path.
3. Final step is finding the state which maximizes $\delta_T(i)$ for $1 \leq i \leq n$.

The first step of calculating the $\delta$ value is to assign each $\delta$ value the probability of starting in each of the states. The recursion step continues to keeps the maximum value throughout the model. The final step finds the final hidden node which is the maximum of all the $\delta_i$'s at time T.

The third and final problem that HMMs should be able to solve is the learning problem. To learn, assume you have a number of observations with known results. These observations are used to calculate new parameters that maximize the probability of the observations given the model, $P(O|\lambda)$. To do this, re-estimation of parameters for the model must increase $P(O|\lambda)$ where the re-estimated model parameters are defined as:

$$\bar{\pi}_i = expected\ number\ of\ times\ in\ state\ i\ at\ time\ t = 1$$

$$\bar{a}_{ij} = \frac{expected\ number\ of\ transitions\ from\ state\ i\ to\ state\ j}{expected\ number\ of\ transitions\ from\ state\ i}$$

$$\bar{b}_{jh} = \frac{expected\ number\ of\ times\ in\ state\ j\ observing\ observation\ V_h}{expected\ number\ of\ times\ in\ state\ j}$$

$$\bar{c}_i(h, k) = \frac{expected\ number\ of\ times\ in\ state\ i\ observing\ V_h\ at\ time\ t\ and\ observing\ V_k\ at\ time\ t+1\ for\ all\ t}{expected\ number\ of\ times\ in\ state\ i\ observing\ V_h}$$

First, define the variable $\gamma_i(t)$ to be the probability of being in state i at time t for the sequence of observations O and the model $\lambda$. That is:

$$\gamma_i(t) = P(Q_t = i | O, \lambda)$$
$$= \frac{P(Q_t = i, O | \lambda)}{P(O | \lambda)}$$
$$= \frac{P(Q_t = i, O | \lambda)}{\sum_{j=1}^{N} P(Q_t = j, O | \lambda)} \quad (eq.1)$$

Earlier, the variable $\alpha_i(t)$ was defined in the forward algorithm as
$\alpha_i(t) = P(O_1, O_2, \ldots, O_t, Q_t = i | \lambda)$
and $\beta_i(t)$ was defined in the backwards algorithm as
$\beta_i(t) = P(O_{t+1}, O_{t+2}, \ldots, O_T | Q_t = i, \lambda)$
Multiplying these together to get
$\alpha_i(t) \cdot \beta_i(t) = P(O_1, O_2, \ldots, O_t, Q_t = i | \lambda) \cdot P(O_{t+1}, O_{t+2}, \ldots, O_T | Q_t = i, \lambda)$
which is the same as $P(Q_t = i, O | \lambda)$. Thus, we can calculate $\gamma_i(t)$ from equation 1 as

$$\gamma_i(t) = \frac{\alpha_i(t) \cdot \beta_i(t)}{\sum_{j=1}^{N} \alpha_j(t) \cdot \beta_j(t)}$$

Now, define the new variable $\xi_{ij}(t)$ to be the probability of being in state i at time t and state j at time t+1 given the observation sequence O and the model $\lambda$. That is
$$\xi_{ij}(t) = P(Q_t = i, Q_{t+1} = j | O, \lambda)$$
The equation for $\xi_{ij}(t)$ can be re-written as
$$\xi_{ij}(t) = \frac{P(Q_t = i, Q_{t+1} = j, O | \lambda)}{P(O | \lambda)}$$
$$= \frac{P(Q_t = i, Q_{t+1} = j, O | \lambda)}{\sum_{i=1}^{N} \sum_{j=1}^{N} P(Q_t = i, Q_{t+1} = j, O | \lambda)} \quad (eq.2)$$

From the forward algorithm, we know that at state $Q_t = i$ and left is equal to $\alpha_i(t)$. From the backwards algorithm, we know that at state $Q_{t+1} = j$ and right is equal to $\beta_j(t)$. We also know that the probability of going from state $Q_t = i$ to $Q_{t+1} = j$ with our observations is equal to (the probability of transition from state i to state j) times (the probability of observing $O_{t+1}$ at state j) times (the probability of being in state i observing $O_t$ and next observing $O_{t+1}$); i.e. $P(Q_t = i, Q_{t+1} = j, O | \lambda) = \alpha_i(t) \cdot a_{ij} b_{jk} c_i(h,k) \cdot \beta_j(t)$. Equation 2 can now be written as:

$$\xi_{ij}(t) = \frac{\alpha_i(t) \cdot a_{ij} b_{jk} c_i(h,k) \cdot \beta_j(t+1)}{\sum_{i=1}^{N} \sum_{j=1}^{N} \alpha_i(t) \cdot a_{ij} b_{jk} c_i(h,k) \cdot \beta_j(t+1)}$$

Notice that if you sum $\gamma_i(t)$ across all of t, $\sum_{t=1}^{T} \gamma_i(t)$, you get the expected number of times in state i. Also, if you sum $\xi_{ij}(t)$ across all of t, $\sum_{t=1}^{T-1} \xi_{ij}(t)$, you get the expected transitions from state i to state j. We can now write the equations for re-estimating the parameters of the model with the given observations to maximize $P(O|\lambda)$.

$$\bar{\pi}_i = expected\ number\ of\ times\ in\ state\ i\ at\ time\ t = 1$$
$$\bar{\pi}_i = \gamma_i(1)$$

$$\bar{a}_{ij} = \frac{expected\ number\ of\ transitions\ from\ state\ i\ to\ state\ j}{expected\ number\ of\ transitions\ from\ state\ i}$$
$$\bar{a}_{ij} = \frac{\sum_{t=1}^{T-1} \xi_{ij}(t)}{\sum_{t=1}^{T-1} \gamma_i(t)}$$

$$\bar{b}_{jh} = \frac{expected\ number\ of\ times\ in\ state\ j\ observing\ observation\ V_h}{expected\ number\ of\ times\ in\ state\ j}$$

$$\bar{b}_{jk} = \frac{\sum_{\substack{t=1 \\ s.t.O_t=V_k}}^{T} \gamma_j(t)}{\sum_{t=1}^{T} \gamma_j(t)}$$

$$\bar{c}_i(h,k) = \frac{\text{expected number of times in state } i \text{ observing } V_h \text{ at time } t \text{ and observing } V_k \text{ at time } t+1 \text{ for all } t}{\text{expected number of times in state } i \text{ observing } V_h}$$

$$\bar{c}_i(h,k) = \frac{\sum_{\substack{t=1 \\ s.t.O_t=V_h, \\ O_{t+1}=V_k}}^{T-1} \gamma_i(t)}{\sum_{\substack{t=1 \\ s.t.O_t=V_h}}^{T-1} \gamma_i(t)}$$

If the current model is $\lambda = (\pi, A, B, C)$ and the re-estimated model is $\bar{\lambda} = (\bar{\pi}, \bar{A}, \bar{B}, \bar{C})$ then either the model $\lambda$ is close enough to $\bar{\lambda}$ that both models can be considered identical at the local critical point, i.e. $\lambda = \bar{\lambda}$, or $P(O|\bar{\lambda}) \geq P(O|\lambda)$, which is proven by Baum [10]. With the new model and the re-estimated parameters, continue to iterate through the observations with their known outcomes until $\lambda = \bar{\lambda}$. At this point, you will be in a local critical point which we will consider the best parameter estimates within this local area.

To prove that the re-estimated equations are correctly derived, we can mathematically optimize the parameters in the likelihood function, P(O|λ), so that it is maximized with the observations with known outputs. This is accomplished using Expectation Maximization (EM) techniques, very similar to [13], which computed the re-estimated parameters of $\bar{\pi}_i$, $\bar{a}_{ij}$, and $\bar{b}_{jh}$. Following the same methods, the Baum Auxiliary Function is defined as $Q(\lambda, \bar{\lambda}) = \sum_{all\ q} [P(Q=q|O=o,\lambda) \cdot \log P(Q=q, O=o|\bar{\lambda})]$. Since

$$P(Q=q|O=o,\lambda) = \frac{P(Q=q, O=q|\lambda)}{P(O=o|\lambda)}$$

and

$$P(Q=q, O=o|\bar{\lambda}) = \bar{\pi}_{q(1)} \prod_{t=1}^{T-1} \bar{a}_{q(t)q(t+1)} \prod_{t=1}^{T} \bar{b}_{q(t)o(t)} \prod_{t=1}^{T-1} \bar{c}_{q(t)}(o(t), o(t+1))$$

the log of this can separate out parts of the above equation:
$$\log P(Q=q, O=q|\bar{\lambda})$$
$$= \log \bar{\pi}_{q(1)} + \sum_{t=1}^{T-1} \log \bar{a}_{q(t)q(t+1)} + \sum_{t=1}^{T} \bar{b}_{q(t)o(t)} + \sum_{t=1}^{T-1} \bar{c}_{q(t)}(o(t), o(t+1))$$

where q(t) and q(t+1) are the states at time t and t+1, and o(t) and o(t+1) are the observation at time t and t+1.

Now the Baum Auxiliary Function can be re-written as

$$Q(\lambda, \bar{\lambda}) = \sum_{all\ q} \left[ \frac{P(Q=q, O=q|\lambda)}{P(O=o|\lambda)} \right] \cdot [\log \bar{\pi}_{q(1)} + \sum_{t=1}^{T-1} \log \bar{a}_{q(t)q(t_1)} + \sum_{t=1}^{T} \bar{b}_{q(t)o(t)}$$
$$+ \sum_{t=1}^{T-1} \bar{c}_{q(t)}(o(t), o(t+0))]$$

Or simply

$$Q(\lambda, \bar{\lambda}) = Q_\pi(\lambda, \bar{\pi}) + Q_a(\lambda, \bar{a}) + Q_b(\lambda, \bar{b}) + Q_c(\lambda, \bar{c})$$

where

$$Q_\pi(\lambda, \bar{\pi}) = \sum_{i=1}^{N} \frac{P(Q=q, O=o|\lambda)}{P(O=o|\lambda)} \log\bar{\pi}_i$$

$$Q_a(\lambda, \bar{a}) = \sum_{i=1}^{N}\sum_{j=1}^{N}\sum_{t=1}^{T-1} \frac{P(q(t)=i, O=o|\lambda)}{P(O=o|\lambda)} \log\bar{a}_{ij}$$

$$Q_b(\lambda, \bar{b}) = \sum_{j=1}^{N}\sum_{k=1}^{M} \sum_{\substack{t=1 \\ st\ o_t = v_k}}^{T} \frac{P(q(t)=j, O=o|\lambda)}{P(O=o|\lambda)} \log\bar{b}_{jk}$$

$$Q_c(\lambda, \bar{c}) = \sum_{i=1}^{N}\sum_{h=1}^{M}\sum_{k=1}^{M} \sum_{\substack{t=1 \\ st\ o_t = V_h \\ o_{t+1} = V_k}}^{T-1} \frac{P(q(t)=i, O=o|\lambda)}{P(O=o|\lambda)} \log\bar{c}_i(o(t), o(t+1)) \quad (eq.3)$$

with the following constraints

$$\sum_{i=1}^{N} \pi_i = 1$$

$$\sum_{j=1}^{N} a_{ij} = 1$$

$$\sum_{k=1}^{M} b_{jk} = 1 \ for\ all\ j$$

$$\sum_{k=1}^{M} c_i(h, k) = 1 \quad (eq.4)$$

Lagrange multipliers can be used to solve for the estimated parameters of $\bar{\lambda} = (\bar{\pi}, \bar{a}, \bar{b}, \bar{c})$. This has been solved of $\bar{\pi}_i$, $\bar{a}_{ij}$, and $\bar{b}_{jh}$ in [13]. To solve for $\bar{c}_i(h, k)$ the same techniques used to solve for the other parameters are used here. Let ρ be the Lagrange multiplier. Solving for $Q_c(\lambda, \bar{c})$, combine the constraint, equation 4, with the Lagrange multiplier and put with equation 3, differentiated and set to 0 for a solution; which is then put into the equations for the final solution of $\bar{c}_i(h, k)$:

$$\frac{\partial}{\partial \bar{c}_i(h,k)}\left[\sum_{i=1}^{N}\sum_{h=1}^{M}\sum_{k=1}^{M}\sum_{\substack{t=1, \\ st\ o_t=V_h \\ o_{t+1}=V_k}}^{T}\left(\frac{P(q(t)=i, O=o|\lambda)}{P(O=o|\lambda)}\log\bar{c}_i(h,k)\right) + \rho\left(\sum_{k=1}^{M} c_i(h,k) - 1\right)\right]$$
$$= 0$$

$$\sum_{\substack{t=1, \\ st\ o_t=V_h, \\ o_{t+1}=V_k}}^{T} \frac{P(q(t)=i, O=o|\lambda)}{P(O=o|\lambda)} \cdot \frac{1}{\bar{c}_i(h,k)} + \rho = 0 \ for\ all\ i, h, k$$

$$\bar{c}_i(h,k) = -\frac{\sum_{\substack{t=1,\\ st\ O_t=V_h,\\ O_{t+1}=V_k}}^{T} \frac{P(q(t)=i, O=o|\lambda)}{P(O=o|\lambda)}}{\rho}$$

Sub this into the constraint equation 4 to get

$$\rho = -\sum_{k=1}^{M} \sum_{\substack{t=1,\\ st\ O_t=V_h\\ O_{t+1}=V_k}}^{T} \frac{P(q(t)=i, O=o|\lambda)}{P(O=o|\lambda)}$$

The right part of this equation can be reduced by combining the summation together to get

$$\rho = -\sum_{\substack{t=1,\\ st\ O_t=V_h}}^{T} \frac{P(q(t)=i, O=o|\lambda)}{P(O=o|\lambda)}$$

Finally, $\bar{c}_i(h,k)$ can be written as

$$\bar{c}_i(h,k) = \frac{\sum_{\substack{t=1,\\ st\ O_t=V_h\\ O_{t+1}=V_k}}^{T} P(q(t)=i, O=o|\lambda)}{\sum_{\substack{t=1,\\ st\ O_t=V_h}}^{T} P(q(t)=i, O=o|\lambda)}$$

which is the same as

$$\bar{c}_i(h,k) = \frac{\sum_{\substack{t=1,\\ st\ O_t=V_h\\ O_{t+1}=V_k}}^{T} \gamma_i(t)}{\sum_{\substack{t=1,\\ st\ O_t=V_h}}^{T} \gamma_i(t)}$$

## 4. EXAMPLE 1: FISHER IRIS DATA OF EXAMPLE OF MESSY DATA

### 4.1 Introduction to the Problem

This chapter will discuss the application of the theory of Evidence Feed Forward HMMs to classify. A great amount of thought on what type of data should be used to focus on the difference in Evidence Feed Forward HMMs and standard HMMs. It has always been a problem with standard HMMs that errors in data give poor classification results. The data provided here will be over-processed to a point where standard techniques to classify this data cannot be accomplished algorithmically or through human inspection. Although there may be viewable patterns in the HMM input data, there will also be errors.

At this point, the reader should come to the conclusion that Evidence Feed Forward HMMs should be pretty good at identifying visual data from people. This is because with Evidence Feed Forward HMMs, the link between observation to observation gives insight into patterns of movement. For example, if a baseball player is pitching a ball, the pitching arm is moving in generally the same direction for most of the action. The linkage from observation to observation will identify continual motion better than standard HMMs, thus more probability

International Journal of Artificial Intelligence & Applications (IJAIA), Vol.2, No.1, January 2011

will be associated with the trained data of pitching to increase probability in the motion of the pitching arm. This will increase likelihood which will provide better results.

However, video and image data of people are not the only category which will benefit from Evidence Feed Forward HMMs. From the simple weather example presented earlier, it should be clear that the linkage of observation to observation will provide a means of increasing probability associated with the observations outside of the hidden states.

The over-processed Fisher Iris data example will demonstrate the increased classification ability of Evidence Feed Forward HMMs in non-video/image data. Although this may look like a stretch with relation of different data sets, it will be shown that there is some linkage and that Evidence Feed Forward HMMs provide a better means of classifying the over-processed data.

The Fisher Iris Data [11] is a classical data set used throughout the image processing and pattern recognition communities as example data and test data for classification of algorithms. R.A. Fisher used this data in 1936 for pattern recognition. It was originally gathered from an American botanist, Edgar Anderson in 1935. The data consists of a collection of sepal width, sepal length, petal width, and petal length of three types of Iris flowers, the Setosa, the Versicolour, and the Virginica. 150 data sets are provided with 50 of each Iris type. One class is linearly separable from the others.

Using this data will help show that the Evidence Feed Forward HMM works better than standard HMMs on more than just image and video data classifications. For this data sets, the information will be over-processed to a point of non recognition visually through the human eyes. That is, there will be no recognizing features when viewing this data visually. The reason for over-processing the data is two-fold. First, HMMs have inherently been bad at classification with messy data. The over-processing of the data will make messy data. Second, this data will show a better example of the increased capabilities of the Evidence Feed Forward HMM when compared with a standard HMM on the same data.

### 4.2 Pre-processing of the Data

The Fisher Iris data was imported from the University of California – Irvine Machine Learning Repository [11] which contains the original data with two fixes as described at the website. These fixes were errors in the data when originally used. A plot of the petal length versus the petal width shows the Setosa class linearly separable from the other two classes, shown in Figure 3.

After importing the data, each category was identified with the highest and lowest values. These values made up the end points for the class category. For example, the petal length ranges from 1.0 cm to 6.9 cm, so the highest value is 6.9 and the lowest value is 1.0. The value range decided the values associated with each bin. A bin, in this case, is a value associated with a range of values from the data. Ten bins were used. For example, Petal Length has a range of 5.9 cm (6.9 cm – 1.0 cm), thus each bin is split 0.59 cm wide. Bin 1 holds the values from 1.0 to 1.59. Bin 2 holds the values of greater than 1.59 to 2.18, and so on. The same holds true for all four categories.

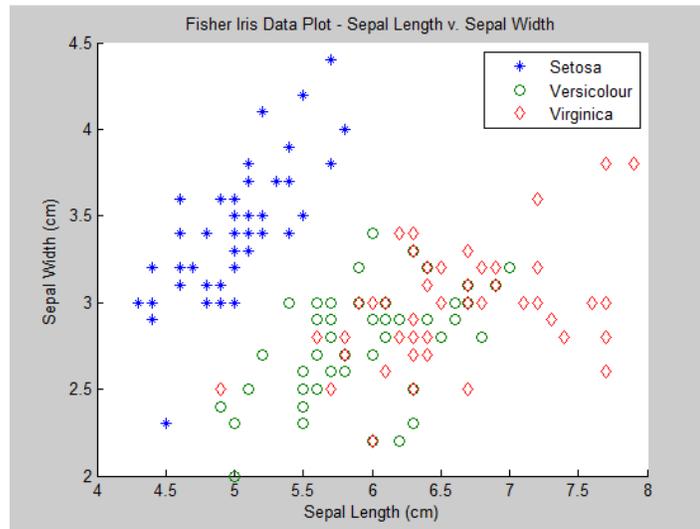

Figure 3: Fisher Iris data plot of petal length versus petal width.

After the bins were created, the new data was processed to find the trends from sepal length to sepal width, sepal width to petal length, and petal length to petal width. That is, increase, decrease, or no change in bin values for each flower. So, if sepal length had a higher bin value then sepal width then this would be viewed as a decrease. If it was a lower bin value then it would be an increase. If the bin values were the same, then no change is recorded. This was done three times (sepal length to sepal width, sepal width to petal length, and petal length to petal width). This is the inputs into the HMMs for comparison.

### 4.3 Results and Analysis

The data used for both training and testing of the Evidence Feed Forward HMM and the standard HMM is described above. Both the Evidence Feed Forward HMM and the standard HMM models were set up to have three hidden nodes and three outputs. More nodes were trained and tested with little to no changes in the results. MatLAB was used to program both the Evidence Feed Forward HMM and the standard HMM. Each HMM was trained with ten data items per flower grouping creating three separate HMMs, one for each Iris type. Training was considered complete when a change in likelihood value of 0.01 or less was attained. This was assumed a good value after several runs through the training and testing. Values that were less than 0.01 took several more iterations to converge. Values higher than 0.01 threshold converged quickly with worse results when testing the data. From the remaining 120 pieces of data both the standard HMM and the Evidence Feed Forward HMM was tested by running the data through each of the three HMMs. The highest likelihood was the winner.

The results showed that for the standard HMM, the likelihood of correctly identifying the Iris category is 35%. This number is very low and can normally be considered a guess since it is about one third correct classification on three possible categories. However, after looking at the results one can see that it correctly categorized almost all the Setosa Iris data but almost none of the other two Iris types. The Setosa Iris is nearly linearly separable in sepal length verse sepal width and in petal length versus petal width. This gives a clue into why the Setosa category was almost completly categorized correctly.

The Evidence Feed Forward HMM did much better in categorizing the over-processed Iris data. It correctly categorized 75% correctly. This is probably because of the linkage between the observations (evidence). Observations are the increase, decrease, or no change to the bins from sepal length to sepal width, sepal width to petal length, and petal length to petal width. There seems to be more of a pattern when a bin value stays the same, increase, or decrease from one bin to another. The bin changes are captured better by the Evidence Feed Forwards HMM than the standard HMM due to the observation to observation transition probability, as shown in the results.

Although these classification rates are still poor when compared to other classifiers, one has to remember the type of data that these are trained and tested on as well as comparing the classifiers within their own category. This data has been over processed and thus, very messy. There are no visual recognizable patterns that stand out in this pre-processed data.

HMMs are notorious for being poor classifiers with messy data. This is shown with the standard HMM on this data, only correctly classifying 35% correctly. However, the Evidence Feed Forward HMM has given much better results, classifying 75% correctly. This better classification rate than standard HMMs is due to the linkage in Evidence Feed Forward Observations which is not present in standard HMMs.

## 5. EXAMPLE 2: VIDEO ACTION RECOGNITION WITH SPARSE MESSY DATA

### 5.1 Introduction to the Problem

The video action recognition data used is from a popular source which created this data to be used to improve people's action classification algorithms on visual data. The database comes from Carnegie Melon University (CMU) Graphics Lab and can be found at [12]. For this example, the simulated data on jumping, running (jogging), dribbling a basketball, and soccer kick was used. Like the Fisher Iris data, this data was over processed to better show the reader the ability to classify with sparse messy data; sparse data because in some cases there where only a handful of action sequences to test and train on.

The CMU data was captured as a sequence of simulated people singled out to perform an activity. Each simulated action is performed different between the groups. The starting and stopping frames are different as is the way the activity is performed. For example, comparing two running sequences, one may take a longer stride than the other or arm movements may have different frequencies.

This data was processed so that the hands, arms, and head can be tracked. It was over processed to a point where one could say it is unusable. This was to show the strength of Evidence Feed Forward HMMs compared to other HMMs. The hands, feet and head location were often misclassified which produced large, abnormal movements with hands, head, and/or feet.

### 5.2 Pre-processing the Data

This data was easier to process then real data since there is no background information that needed to be segmented out or tracking of colors. Like many image processing exercises, the more processing that is done to an image, the more likely that the image is no longer a good representative of the items you are trying to find. In this case, we are trying to find links between frames to represent actions.

The determination of the edges of the body was by threshold analysis. If the grayscale image of the body was below a threshold, then it was considered not to be part of the body. As images are taken, there is usually a mixing of colors from the edges of the body to the background. This mixture is removed using thresholding. If thresholding was not used then the edges of the image would be very rough and virtual impossible to perform some of our data processing techniques to it. The thresholding produced a binary image.

Next the binary image of each frame was skeletonized. That is, the area representing the image was thinned to a point where only a single line represented the body parts. This made it easier to identify the tips of the hands, feet, and head through the endpoints of skeletal image frame. If a point in the image frame had only one neighbor, then it was considered an endpoint. Often times, the skeletonization process thins the image to much where several end points other than the tips of the hands, feet, and head are found. This creates extra points in the image.

To reduce the number of extra points, another process is performed to find the location of the last points and keep all points that are within a certain distance from them. This distance must be greater than the pre-determined distance otherwise it is considered part of the point. This alleviates some of the problems with multiple points, but not all of them. So, only the five points which best represent the last five points are kept.

Looping of an endpoint occurs from the thresholding process, which the skeletonization process exploits. When a loop occurs, there is no endpoint since all points in the looped area have more than one neighbor. This eliminates key endpoints. To add the end points back in, a close representative of the point is used. It keeps the location of the last point which was missing and assumes the change is negligible. As one can imagine, this may cause some problems with the classification since changes are important with Evidence Feed Forward HMMs. When automating the pre-processing phase of the image sequences, this problem occurs often, so no measures are made to manually reduce it. The idea is to have some uniformity in the pre-processing as well as the classification.

The worst of these errors happen when a point is missing in one frame and then shows up as extra points in the others. Often times the skeletonization process produces lines where none should exist and creates loops where there should exist a single line, all in the same image frame. In this case, it may have found the five points it was looking for, but not the correct ones. This error is propagated into the classification algorithm as well as in the pre-processing procedure since it uses the previous frame as a template to where the points should be.

Without all the image processing errors, this would be a viable way to find activities of the people. The hands, feet, and head location would probably be tracked and with enough data, the classification algorithm would utilize all the position information. However, at this point it should be noted that first, this is not an image processing research project. The errors in the pre-processing are common and with a greater amount of time many may be alleviated. However, to demonstrate the robustness due to errors in Evidence Feed Forward HMMs versus standard HMMs, these types of errors will not be removed.

To continue along with the pre-processing of this data, the five points are feed into another processing step which is meant to find and track the changes in the position of the points in relation to each other. The five points will make up a bounding box which all five points are either on the edge of the box or inside. Bounding boxes are often used in pedestrian detection to encompass the entire pedestrian within the box. This process is similar except that it only encompasses the hands, feet, and head. If you can imagine a person standing sideways with hands out, the box would be thinner than if a person is standing facing forward with hands to their sides and feet spread apart. These differences are what we are looking for.

The height/width ratio is found by taking the height and dividing by the width. This method is sometimes used as a simple way of determining body position from crouching to standing, however in this case, only hands, feet, and head are used to create the bounding box where usually the entire body is used to create the bounding box.

The height/width ratio for each frame in the sequence of images is put into one of eleven groups based on the value of the ratio. Group 1 is if the ratio is below 0.1, group 2 is for the ratio to be between 0.1 and 0.2, and so on to group 11 which is above 1.0. It is assumed that the hands, feet, and head would be changing throughout the activity and these changes can be tracked to produce a similar activity representation per individual. In reality, this simplifies the activity and would be true if this was applied to a single person's way of performing the activity. However, everyone has a relatively unique body position at each stage of the activity. This over simplification of the activity is part of the over-processing which is often a problem with many classifiers.

The final step of pre-processing is to track and record these changes in groups from frame to frame. If a frame changes to a higher group then the previous frame then it is considered an INCREASE, if it is lower than it is a DECREASE. If there is no change in grouping then it is considered NO CHANGE. This creates a way of tracking the changes from frame to frame for input into the HMMs. However, this process does not track the magnitude of the change which might also be important. The idea is to keep the data simple and represent the output in our HMMs. If the observation data was more complicated, i.e. more values associated with it, then the training data would need to be increased to cover all the possible options of the observation data.

The pre-processed data represents the evidence, or observations, for our HMMs. The evidence is the input into the HMMs for both training and testing. The number of frames per activity varies greatly, anywhere from 20 to 120 frames per activity. it was decided that the likelihood of each testing activity would be compared and the greatest likelihood would be the class it represented.

## 5.3 Results and Analysis

Of the four classes, there are a total of 64 activities modeled by the simulated sequences. 17 of these activities are jumping, 28 are jogging/running, 7 are soccer kicking, and 12 are dribbling a basketball. The dribbling a basket ball had 9 from the right hand and 3 from the left. From each group, except for soccer kicking, four
random activities were used as training the Evidence Feed Forward HMM and the standard HMM. Three were taken for soccer kicking because the sample was very small. These training activities were the same for training both HMMs. The training and testing of the Evidence Feed Forward HMM and the standard HMM were programmed in MatLAB.

A total of 3 nodes were used in the hidden layer. More were tried, but did not increase the classification rate. With less nodes, the classification rates were worse, so three nodes were optimal on our testing/training samples. Also, from earlier, three input values are used as the evidence which roughly represented the change in body position from frame to frame (INCREASE, DECREASE, and NO CHANGE). The training was considered complete when the threshold changes of the previous HMM log-likelihood and the new HMM log-likelihood with updated parameters were less than 0.25. Many different thresholds where tried with 0.25 being the optimal one. With a lower threshold, the training took a long time to converge on a solution for many of the activities. This usually means over-training and as a result the testing data was classified correctly at a worse rate than with a threshold of 0.25. When increasing the threshold above 0.25, the training converged very quickly and the results were also not as good when the testing data was applied. This is indicative of under training. A threshold change value of 0.25 was considered optimal.

For the standard Hidden Markov Model, it correctly classified 67% of the jump activity, 21% of the jogging activity, 50% of the soccer kick activity, and 12.5% of the dribbling activity. See Table 1 below. The Evidence Feed Forward HMM correctly identified 78% of the jump activity, 46% of the jogging activity, 100% of the soccer kick activity, and 50% of the dribbling activity.

Notice that in all cases, the Evidence Feed Forward HMM did better than the standard HMM on this data. This is because, even though the data is sparse and it was over processed, the Evidence Feed Forward HMM was still able to pick up some of the patterns that recognized the activity due to its ability to increase predictions based on previous observations. That is, the observation to observation link in the Evidence Feed Forward HMM provides the ability to increase the overall probability of an activity's pattern, and thus increase the classification rate.

Table 1.
Classification rates of HMM and Evidence Feed Forward HMM.

| Activity | HMM | EFF-HMM |
| --- | --- | --- |
| Jump | 67 | 78 |
| Jogging | 21 | 46 |
| Soccer kick | 50 | 100 |
| Dribbling a basketball | 12.5 | 50 |

To further improve on the analysis one must first look at how the data is perceived to the classifier. For the jumping data, one can imagine the ratio starting small, getting bigger as the person crouches to jump, then getting small again when the person stretch through the jump. The patterns of the groupings should be decreasing for a while then increasing. For running, the arms stretch out around the time the legs stretch out and they come together as the legs come together. Therefore, one would expect, depending on the starting phase of the sequence simulating running, the bounding box ratio would be small (arms and legs together) then increasing until a certain point (when the arms and legs are at their peak) then decreasing. This is similar to the jumping and it is noted that many of the misclassified activities of jogging were classified as jumping.

For the dribbling a basketball activity, one would expect that the bounding boxes would be pretty much the same throughout. As the person walks, the dribbling hand continues up and down at the same position but with further analysis, it shows that some of the simulations move their opposite arm similar to walking which makes the bounding box increase and decrease, thus the changes in the bounding box ratio will also increase and decrease in a similar fashion to jogging and jumping.

Soccer kicking was a bit different than the rest of the activities. As the player lines up towards the ball the legs come together. Further into the sequence the right arm extends out as the kicking process is started. Next the leg extends out. Most of the bounding box ratio from frame to frame is increasing. This is easy for the soccer kicking to classify correctly. However, some of the Jogging activity also was classified as soccer kicking. This is because the way the running simulation in these four people had a large amount of increasing which better associated with soccer kicking then with running.

These results not only shows why some of the activities were misclassified, but also shows the problem with over-processed data. Since there are no real distinguishing features that are indicative to the activity, many of the activities were classified with similar events as far as their data representation is concerned.

The Evidence Feed Forward HMM did much better than the standard HMM in this sparse, over-processed data. This is due to the linkage of observations to observations. Even though there were no visible patterns associated with the activity, the Evidence Feed Forward HMM showed that fairly good classification could still be accomplished when compared to standard HMM classification.

## 6. CONCLUSION

The Evidence Feed Forward Hidden Markov Model is more than a standard Hidden Markov Model. It provides observation to observation linkage in the algorithm. The linkages were developed through analysis and proven mathematically. It was originally designed for better classification of visual action data, like the differences in pitching and throwing from the outfield. The idea is that if there existed a way to link one observation frame to another, then there may be some patterns that the Evidence Feed Forward HMM could recognize better than if there were no observation linkages, like how a standard HMM would classify. This was extended for more than just visual data and has shown to work in other classification areas.

The proposed research on the development of the Evidence Feed Forward HMM has worked well for classifications of items based on the observation to observation link that is not available in other types of HMMs. With messy data, it has outperformed the standard HMM in classification of the Iris data. With sparse messy data it has outperformed classification on visual activities created by simulated actions from the CMU data set. This dataset was over-processed to a point where no visual cues were seen.

The mathematical proofs, results, and analysis of the Evidence Feed Forward HMM has added a new type of classifier and a new way of thinking about how the data should be viewed. Observation to observation linkages are essential, especially for visual data, if a more complete classification system is to be developed.